\documentclass[10pt,twocolumn,letterpaper]{article}

\pdfoutput=1

\usepackage{cvpr}
\usepackage{times}
\usepackage{epsfig}
\usepackage{graphicx}
\usepackage{amsmath}
\usepackage{amssymb}
\usepackage[title]{appendix}
\usepackage{placeins} 
\usepackage{dblfloatfix} 
\usepackage{subcaption} 
\usepackage{enumitem} 
\usepackage{authblk}

\usepackage{listings} 

\usepackage[pagebackref=true,breaklinks=true,letterpaper=true,colorlinks,bookmarks=false]{hyperref}

\cvprfinalcopy 

\def\cvprPaperID{****} 

\begin{document}

\title{SMILER: Saliency Model Implementation Library for Experimental Research}

\author[1]{Calden Wloka}
\author[1]{Toni Kuni{\'c}}
\author[1]{Iuliia Kotseruba}
\author[2]{Ramin Fahimi}
\author[3]{Nicholas Frosst}
\author[4]{Neil D.B. Bruce}
\author[1]{John K. Tsotsos}

\affil[1]{Department of Electrical Engineering and Computer Science\\
  York University, Toronto, Canada}
\affil[2]{Department of Computer Science\\
University of Manitoba, Winnipeg, Canada}
\affil[3]{Google Brain\\
Toronto, Canada}
\affil[4]{
Department of Computer Science\\
Ryerson University, Toronto, Canada}

\affil[ ]{
{\vskip 0em{\tt\small $^1$\{calden, tk, yulia, tsotsos\}@eecs.yorku.ca\\
$^2$ramin@cs.umanitoba.ca\\
$^3$frosst@google.com\\
$^4$bruce@ryerson.ca}}}


\maketitle

\begin{abstract}

The Saliency Model Implementation Library for Experimental Research (SMILER) is a new software package which provides an open, standardized, and extensible framework for maintaining and executing computational saliency models. This work drastically reduces the human effort required to apply saliency algorithms to new tasks and datasets, while also ensuring consistency and procedural correctness for results and conclusions produced by different parties. At its launch SMILER already includes twenty three saliency models (fourteen models based in MATLAB and nine supported through containerization), and the open design of SMILER encourages this number to grow with future contributions from the community. The project may be downloaded and contributed to through its GitHub page: \url{https://github.com/tsotsoslab/smiler}

\end{abstract}

\section{Introduction}
\label{sec:Intro}

Many aspects of modern scientific research are heavily dependent on software. This dependence raises a number of challenges, including the fact that software developed primarily for research is often difficult or time consuming to set up and execute, and may include undocumented assumptions, parameters, or conflicting requirements which present a major impediment to research sharing and reproducibility \cite{Goble2014,AlnoamanyBorghi2018}. The field of saliency research is an area in which many of these challenges may be seen: over the past two decades there has been a dramatic increase in both the number and nature of computational saliency models. Not only does this volume make it increasingly difficult for researchers to effectively explore and test the landscape of different approaches to saliency modeling, the lack of a standard interface to each model increases the likelihood that any given model may be incorrectly or erroneously configured, leading to mistaken or inconsistent results in the saliency literature.

For example, Table~\ref{tab:Inconsistent} shows scores computed by the similarity (SIM) metric~\cite{JuddEtAl2012} as calculated in three different studies (Vig~\etal~\cite{VigEtAl2014}, Wang and Shen~\cite{WangShen2018}, and Berga and Otazu~\cite{BergaOtazu2018}) on the Toronto dataset \cite{BruceTsotsos2006}. Note that not only do the scores not match for even a single algorithm across any two studies, but also the rank order of performance shifts. For instance, Vig~\etal find that their eDN model~\cite{VigEtAl2014} outperforms the CAS model~\cite{GofermanEtAl2012} both with and without added center bias, whereas Wang and Shen find that CAS outperforms eDN. Similarly, Vig~\etal find that with center bias added, AWS~\cite{GarciaEtAl2012} outperforms AIM~\cite{BruceTsotsos2006} which outperforms GBVS~\cite{HarelEtAl2007}, and without center bias added the order shifts from best to worst to be GBVS, AIM, AWS. Neither ordering, however, agrees with the results of Berga and Otazu, who find a ranking of best to worst for these three models to be GBVS, AWS, and then AIM.

Note that we are not accusing any authors of impropriety or misconduct, but rather are simply highlighting that without standardization three different studies give rise to three different sets of scores and rankings. This may be further explained through an example of parameter handling: several saliency algorithms have been shown to operate best over colour spaces alternative to RGB, including the Covariance-based Saliency (CVS) model \cite{ErdemErdem2013}, the Image Signature (IMSIG) model \cite{HouEtAl2012}, and the Saliency Detection by Self-Resemblance (SSR) model \cite{SeoMilanfar2009} which all operate best in the CIELAB colour space, and the Quaternion-based Spectral Saliency (QSS) model \cite{SchauerteStiefelhagen2012} which performs best using YUV colour. However, the original model code released by each method's authors handles image input completely differently. CVS expects as input a string argument specifying an image path, then loads the image and converts it to CIELAB space internally. SSR expects as input an image variable in RGB format, which is converted to CIELAB space internally. IMSIG expects as input an image variable in RGB format, which is converted to CIELAB or DKL colour space when provided with an optional parameter setting (when no parameter is provided, IMSIG will process the image in RGB, despite the recommendation of the authors to use CIELAB space). QSS expects an image variable as input, but provides no internal image conversions; the model authors recommend that YUV format be used, and the conversion is expected to be performed by the user before calling the QSS model code. No one approach is any more correct than any other, but this lack of standardization places a non-negligible burden on users and can easily lead to errors or oversights in which a user believes they have configured the models to operate in the desired colour spaces but some subset of models is actually not being applied as expected. When one takes into account the number of additional parameters which must be controlled (such as numerical scaling of saliency map output, post-processing smoothing, the application of a center prior, or any model-specific settings), the burden of use and chance for error is only further compounded. Likewise, for any hope of reproducibility, these parameters must all be exhaustively documented (which, unfortunately, has not always been the case within the literature).

Recent years have also seen a shift in model development toward methods which rely on deep learning networks. While many of these methods achieve very high benchmark performance, they also introduce a new practical challenge for the dissemination and sharing of code. In order to operate in a reasonable timeframe, most deep learning algorithms require a significant share of their computation to take place on a graphical processing unit (GPU). This necessitates that a user not only has access to GPU hardware, but also has the appropriate libraries installed which will allow access to the GPU for calculations. Unfortunately, the setup processes of different GPU scientific computing libraries as part of the same development environment are often fairly involved and daunting for non-experts. Likewise, there is a lack of standardization, and frequently the libraries necessary to run one model will be incompatible with the libraries required for another model. For example, oSALICON \cite{Thomas2016} is implemented in Caffe~\cite{jia2014caffe}, while DeepGaze II \cite{KummererEtAl2016} is implemented in TensorFlow~\cite{tensorflow2015}; as of the time of this writing, following the official setup documentation for one project interferes with the setup of the other. While running both libraries on the same system is possible, it requires knowledge that goes beyond the official documentation. Therefore, not only will the installation of even a single model likely be a large barrier of entry for a user who is not actively pursuing work with deep learning-based development, but also providing simultaneous access to a general-purpose library of saliency models is extremely difficult without isolating incompatible model dependencies from each other. Due to potentially frail assumptions regarding backward compatibility, there is a significant risk that important contributions may be lost to time or not explored in sufficient detail owing to the need to operate within a specific ecosystem.

Our work aims to facilitate research efforts in computational salience by addressing these software challenges. We do so by introducing the Saliency Model Implementation Library for Experimental Research (SMILER). SMILER provides library-like functionality for saliency models, standardizing the input, output, and parameter specifications for each model, and isolating incompatible model components from each other. At the time of this publication SMILER supports twenty three models: Attention by Information Maximization (AIM)~\cite{BruceTsotsos2006}, Adaptive Whitening Saliency (AWS)~\cite{GarciaEtAl2012}, Boolean Map Saliency (BMS)~\cite{ZhangSclaroff2016}, Context Aware Saliency (CAS)~\cite{GofermanEtAl2012} based an open implementation~\cite{TsaiChangCode}, Covariance-based Saliency (CVS)~\cite{ErdemErdem2013}, DeepGaze II (DGII)~\cite{KummererEtAl2016}, Deep Visual Attention Prediction (DVAP)~\cite{WangShen2018}, Dynamic Visual Attention (DVA)~\cite{HouZhang2008}, Ensemble of Deep Networks (eDN)~\cite{VigEtAl2014}, Fast and Efficient Saliency Detection (FES)~\cite{TavakoliEtAl2011}, Graph-based Visual Saliency (GBVS)~\cite{HarelEtAl2007}, Intensity Contrast Features (ICF)~\cite{KummererEtAl2017}, the Itti-Koch-Niebur Saliency Model (IKN)~\cite{IttiKochNiebur1998}, Image Signature (IMSIG)~\cite{HouEtAl2012}, Learning Discriminative Subspaces (LDS)~\cite{FangEtAl2017}, a Deep Multi-Level Network (MLNet)~\cite{CorniaEtAl2016a}, an open implementation~\cite{Thomas2016} of Saliency in Context~\cite{HuangEtAl2015} (oSALICON), Quaternion-Based Spectral Saliency (QSS)~\cite{SchauerteStiefelhagen2012}, RARE2012~\cite{RicheEtAl2013}, Saliency Attentive Model (SAM)~\cite{CorniaEtAl2016}, Saliency Detection by Self-Resemblance (SSR)~\cite{SeoMilanfar2009}, Saliency using Generative Adversarial Networks (SalGAN)~\cite{PanEtAl2017}, and Saliency Using Natural statistics (SUN)~\cite{ZhangEtAl2008}. This set provides a broad representative sample of models popular in the saliency research community focused on fixation prediction, and the system is designed to be easily extensible with additional models.

\begin{table}[!bt]
    \begin{center}
    \begin{tabular}{| c || c | c | c |}
        \hline
        Reference & \cite{VigEtAl2014} & \cite{WangShen2018} & \cite{BergaOtazu2018} \\ \hline \hline
        eDN \cite{VigEtAl2014} & 0.573/0.487 & 0.40 & - \\ \hline
        CAS \cite{GofermanEtAl2012} & 0.555/0.427 & 0.44 & - \\ \hline
        AWS \cite{GarciaEtAl2012} & 0.558/0.407 & - & 0.352 \\ \hline
        GBVS \cite{HarelEtAl2007} & 0.534/0.496 & 0.49 & 0.397 \\ \hline
        AIM \cite{BruceTsotsos2006} & 0.549/0.426 & 0.36 & 0.314 \\ \hline
        IKN \cite{IttiKochNiebur1998} & - & 0.45 & 0.366 \\
        \hline
    \end{tabular}
    \caption{An example of inconsistent model results. Here we show the SIM \cite{JuddEtAl2012} scores for five algorithms over the Toronto dataset \cite{BruceTsotsos2006} as reported by three recent publications (note that \cite{VigEtAl2014} report two scores, one with added center bias and one without). - indicates that a model was not run in that particular study. Note that we are not claiming any wrong-doing on the parts of these studies, but rather pointing out that each study likely executed these models in slightly different ways, leading to inconsistent results and a substantial challenge for reproducibility in the literature. \label{tab:Inconsistent}}
    \vspace{-7mm}
    \end{center}
\end{table}

\subsection{Related Work}
\label{subsec:Background}

The rapid expansion in saliency model numbers has been met in at least one area by a concerted effort at consolidation and standardization: performance benchmarking. Starting with a number of isolated benchmark surveys (\eg see \cite{BorjiItti2012, BorjiEtAl2013b, JuddEtAl2012, RicheEtAl2013}), this effort eventually culminated with the establishment of the MIT Saliency Benchmark \cite{BylinskiEtAl}, a continually updated ranking of saliency algorithm performance over a pair of curated benchmark datasets.

While these benchmarking efforts have provided an important overview of progress in the field of fixation prediction and an impartial ranking of models, the scope of this effort has remained predominantly focused on comparative performance evaluation. The MIT Saliency Benchmark does helpfully provide an index of links to code which has been released by model authors, but the onus of handling setup and operation of each different model's code remains with the user. SMILER, therefore, provides a complementary service to the saliency community; rather than focus on standardizing evaluations of performance, SMILER seeks to standardize the execution of model code and thereby enable exploration of additional research avenues not encapsulated by current benchmarks.

It should be noted that the current collection of models supported by SMILER consists of models which focus on pixel-wise assignment of conspicuity values and which have been predominantly applied to the domain of human fixation prediction. There are, however, other branches of saliency model research, such as salient object detection (\eg see~\cite{ChangEtAl2011} for an early example, and~\cite{HanEtAl2018} for an overview and recent survey). Likewise, the models included are predominantly focused on saliency prediction over static scenes, but there is nevertheless significant interest in saliency over dynamic stimuli (\eg see \cite{MaratEtAl2009,MahadevanVasconcelos2010,ZaharescuWildes2012}). This focus on models which are more representative of fixation prediction over static images is not intended to dismiss or ignore these other research avenues, but rather is meant to form a solid base for the SMILER platform.

\subsection{Our Contributions}
\label{sec:Contributions}

SMILER provides two primary benefits to the saliency research community: reducing the burden of use for code execution, and promoting the consistency and reproducibility of experimental results. The first step in achieving these goals is the establishment of a common model-independent application programming interface (API). In order to make this API as effective as possible in facilitating a wide range of research, we ensure the following qualities:
\begin{itemize}
  \item It should be possible to run each algorithm in a default mode which requires minimal user input or selection of settings, providing an intuitive mode which can be used without expert-level algorithmic familiarity. At the most basic level of function, a model should expect only that an input image is specified, and it should return as output a saliency map corresponding to that image. By default, this saliency map should be the same height and width as the input image.
  \item As much as possible, there should be no loss in the flexibility of parameter options available for each individual model. While it is not possible to have a common set of parameters for each algorithm, to reduce the complexity of operation as much as possible it should be possible to selectively choose which parameters to manually specify, with unspecified parameters automatically populated with default values (thereby allowing for a smooth transition from fully default mode through to fully user-specified operations).
\end{itemize}

By creating a standard interface for model execution, we allow users to learn a single API rather than one for each model. The flexible method for parameter specification allows researchers to engage with models at a variety of levels of depth, from novel benchmarking work using default settings through to the analysis of model behaviour over a range of parameter settings.

By standardizing model execution, we also ensure that when researchers run a given model with particular settings, they are sure to get the same results as when another researcher runs the same model with the same settings. If both researchers were expected to independently set up and execute the model using their own custom scripts, it is entirely possible for unintentional bugs or oversights to lead to inconsistencies between them. Of course, it is entirely possible for SMILER to contain bugs, but by fostering an open and centralized repository for saliency model code, we ensure that when bugs are found and corrected this correction is distributed to all users.

With a straightforward and flexible code base for easily executing a large ecosystem of saliency models, we envision a number of research directions which SMILER can support, including but not limited to:
\begin{itemize}
  \setlength{\itemsep}{0cm}
  \setlength{\parskip}{0cm}
  \item Performance benchmarking on applications outside of fixation prediction for which saliency may be applicable, but extensive performance testing is not currently available. Examples include:
  \begin{itemize}
    \setlength{\itemsep}{0cm}
    \setlength{\parskip}{0cm}
    \item Anisotropic image or video compression (\eg \cite{DhavaleItti2003,YuLisin2009,GuoZhang2010,HardingRoberston2009})
    \item Defect detection (\eg \cite{MancasEtAl2007,BoimanIrani2007})
    \item Image cropping (\eg \cite{SuhEtAl2003}) or retargeting (\eg \cite{ZhuEtAl2011})
    \item Image domains outside the natural images which form the bulk of fixation datasets, such as websites \cite{MasciocchiStill2013} or satellite imagery \cite{FuEtAl2018}
    \item Image quality assessment (\eg \cite{MaZhang2008,YubingEtAl2010,LinEtAl2013})
    \item Robotic navigation (\eg \cite{ChangItti2010,RobertsEtAl2012}) or search (\eg \cite{RasouliTsotsos2014})
  \end{itemize}
  \item Saliency model evaluation for other attentional aspects beyond fixation prediction, such as the psychophysical evaluations proposed in Bruce~\etal~\cite{BruceEtAl2015}.
  \item Increasing the robustness of conclusions for research which compares experimental findings in psychology or neuroscience to saliency algorithms (\eg \cite{NuthmannHenderson2010,HendersonEtAl2009,WhiteEtAl2017}) by allowing comparison against many saliency models rather than a single one.
\end{itemize}

\section{Design Overview}
\label{sec:Design}

In order to leverage as wide a range of existing saliency model implementations as possible, as well as to support researchers with different degrees of computational and software resources available to them, SMILER is comprised of two major programming language components: a MATLAB component and a command-line interface (CLI) implemented in Python. The MATLAB component comes with a subset of available models and is fully cross-platform so long as the computer supports the MATLAB environment and the user has access to the appropriate licenses for MATLAB and any specific toolboxes required by a given model. The CLI is currently only supported for the Linux operating system, but provides access to the full suite of SMILER models, both MATLAB and deep learning. In order to foster open software development and move away from proprietary software systems, the CLI will be the primary focus of future development for the SMILER project, with an emphasis on adding models that do not depend on a MATLAB license. To minimize code drift across multiple interfaces, all models included in SMILER contain a configuration and information file described in Section~\ref{subsec:Config_Files}.

Prior to the shift in algorithm development toward deep learning models, the majority of saliency models were released for the MATLAB programming environment. As a consequence, early development of SMILER was also based in MATLAB. However, the need to handle deep learning models which are predominantly implemented in languages other than MATLAB necessitated a shift to another language. Nevertheless, it was felt that it would not be desirable to drop the MATLAB specific structure which is already in place, as there are many users who would prefer to operate within the MATLAB environment (for example, many researchers are already familiar with MATLAB through the use of tools such as the PsychToolbox \cite{Brainard1997}, and may prefer to keep their research efforts in the same programming environment). Therefore, the design of SMILER retains MATLAB functionality for all algorithms available in the MATLAB environment, as well as a functional MATLAB interface for executing these models. The SMILER CLI utilizes the MATLAB's Python API to allow invocation of MATLAB models in the background, without using the full MATLAB graphical user interface.

Whether one is working through MATLAB or the SMILER CLI, the general principles of SMILER operation remain the same, and the details of operation are kept as close as possible given the different nature of the MATLAB Integrated Development Environment (IDE) and the CLI. An overview of operation for the MATLAB interface is given in Section~\ref{subsec:MATLAB_interface}, and for the SMILER CLI in Section~\ref{subsec:SMILER_CLI}. Due to the more extensive support of saliency models and support for YAML-based experiment specification (discussed more thoroughly in Section~\ref{subsec:SMILER_CLI}), we would encourage users to preferentially use the CLI.

SMILER attempts as closely as possible to maintain the originally intended functionality of each model. However, there are times when this is not possible. For example, although expecting the output map to be the same height and width as the input image seems like a straightforward assumption, it is not the default behaviour of all algorithms. Some models automatically resize input images to a specified size, and return this size as output, whereas others such as SUN \cite{ZhangEtAl2008} make a point of returning only the portion of the image for which output is valid without image padding (trimming the half-width of the feature kernels from the image border). Although this inconsistency between input and output size may be a distinct choice by the model designers with a clear justification, for the purposes of SMILER it was felt that ensuring a common behaviour across algorithms was the more important consideration and therefore SMILER will re-scale or pad as appropriate the saliency maps to be the same dimensions as the original input image.

Models for which the full source code has been released are preferred candidates for inclusion in SMILER, as this allows for more robust crowd-sourced bug checking, access to the full range of algorithm parameters (particularly for post-processing steps such as smoothing), and aids in future code maintenance (for example, the use of deprecated functions which are no longer supported by MATLAB or third party libraries). It should be noted, however, that several models are nevertheless included despite only having access to a pre-compiled version, namely AWS~\cite{GarciaEtAl2012}, FES~\cite{TavakoliEtAl2011}, and RARE2012~\cite{RicheEtAl2013}. The pre-compiled version of CAS~\cite{GofermanEtAl2012} provided by the original study authors is not compatible with SMILER, and therefore an open source implementation \cite{TsaiChangCode} has been used. In a similar vein, code for the SALICON model~\cite{HuangEtAl2015} is not available at the time of this writing, but we include the oSALICON~\cite{Thomas2016} implementation which is based on the original model.

\subsection{A Common Format for Information and Configuration}
\label{subsec:Config_Files}

There are a number of parameters and controls for pre- and post-processing of saliency maps which are common to many or all models. As well, each model in SMILER requires several important attributes to be associated with it, including citation information and model-specific parameters. In order to provide this information in a manner which is extensible to the inclusion of future model properties or specifications and independent of the specific programming interface accessing the model, several JavaScript Object Notation (JSON) configuration files are used. JSON is an open standard for providing human-readable attribute-value pairs, and provides an effective format for storing model information in SMILER.

\begin{lstlisting}[label={lst:config.json},caption={An example global parameter specification from the dictionary contained in \texttt{config.json}.}]
"do_smoothing": {
  "default": "default",
  "description": "Specification for post-processing smoothing.
    'default' uses whatever smoothing step is provided by the originally released model code.
    'none' turns off post-processing smoothing (though it should be noted that some implicit smoothing (eg. through image resizing) will remain).
    'custom' smooths the map with a specified kernel.
    'proportional' smooths the map with a kernel sized to the major dimension of the image.",
  "valid_values": ["default", "none", "custom", "proportional"]
}
\end{lstlisting}

Parameters which affect the execution of a majority of SMILER models are referred to as \emph{global} and are described in a \texttt{config.json} file in the root of the SMILER directory. These parameters and their default values are shown in Table~\ref{tab:default_params}. Listing~\ref{lst:config.json} shows an example JSON parameter specification. The user shouldn't need to modify these JSON files directly, as they contain specifications for the SMILER system. The user should specify parameters at run-time via the MATLAB interface or with a YAML experiment file passed to the SMILER CLI.

\begin{table*}[t]
\centering
\begin{tabular}{|l|l|p{4.9cm}|p{6.7cm}|}
\hline
\textbf{Parameter} & \textbf{Default Value} & \textbf{Valid Values} & \textbf{Description} \\ \hline
do\_smoothing & ``default'' & ``default'', ``none'', ``custom'', ``proportional'' & Specification for post-processing smoothing. \\ \hline
smooth\_size  & 9           & Integer greater than 0. &  Custom smoothing kernel size, only used when do\_smoothing is set to custom.                    \\ \hline
smooth\_std   & 3.0         & Float greater than 0.   &  Custom smoothing kernel standard deviation, only used when do\_smoothing is set to custom.                    \\ \hline
smooth\_prop  & 0.05        & Float greater than 0.   &   Proportional smoothing kernel parameter, only used when do\_smoothing is set to proportional.                   \\ \hline
scale\_output & ``min-max'' & ``min-max'', ``none'', ``normalized''  & Specification for rescaling saliency map values into a specified range. \\ \hline
scale\_min    & 0.0         & Float less than scale\_max.                    &  Minimum saliency value in the map, only used when scale\_output is set to min-max.                    \\ \hline
scale\_max    & 1.0         & Float greater than scale\_min.                 &  Maximum saliency value in the map, only used when scale\_output is set to min-max                    \\ \hline
color\_space  & ``default'' & ``RGB'', ``gray'', ``YCbCr'', ``LAB'', ``HSV'' &  Specification for pre-processing conversion of the image color channels. \\ \hline
\end{tabular}
\caption{Default values for global SMILER parameters, as defined in \texttt{config.json}.
  \label{tab:default_params}}
\end{table*}

As can be seen, \texttt{parameters} are defined as a nested dictionary. Each parameter is populated with three fields: \texttt{default}, \texttt{description}, and \texttt{valid\_values}. The \texttt{default} field is used when no other source of parameter specification is available. The \texttt{description} and \texttt{valid\_values} fields are intended for human consumption; each SMILER interface provides a method for accessing and displaying this information to a user (detailed in Section~\ref{subsec:MATLAB_interface} for the MATLAB interface and Section~\ref{subsec:SMILER_CLI} for the CLI). The \texttt{description} field provides a brief explanation for the role the parameter plays in the calculation of a saliency map, while the \texttt{valid\_values} field provides either an explicit set of available parameter assignments (\eg for the \texttt{scale\_output} parameter there are three options: \texttt{min-max}, \texttt{none}, or \texttt{normalized}) or a specified range (\eg an ``Integer greater than 0'' for \texttt{smooth\_size}).

\begin{lstlisting}[label={lst:AIM_smiler.json},caption={An example \texttt{smiler.json} file showing model-specific information for the AIM algorithm. \cite{BruceTsotsos2006}}]
{
  "name": "AIM",
  "long_name": "Attention by Information Maximization",
  "version": "1.0.0",
  "citation": "N.D.B. Bruce and J.K. Tsotsos (2006). Saliency Based on Information Maximization. Proc. Neural Information Processing Systems (NIPS)",
  "model_type": "matlab",
  "model_files": [],
  "parameters": {
    "AIM_filters": {
      "default": "21jade950.mat",
      "description": "The feature filter set to be used by the AIM algorithm. In the form [size][name][info], where each filter is size by size in dimension, name is the ICA algorithm used to derive the filters, and info provides a measure of the retained information (higher numbers correspond to more filters).",
        "valid_values": [
          "21infomax[900,950,975,990,995,999].mat",
          "21jade950.mat",
          "31infomax[950,975,990].mat",
          "31jade[900,950].mat"
        ]
    }
  }
}
\end{lstlisting}

Each model includes additional model-specific information in a \texttt{smiler.json} file included in the root of its subfolder. An example \texttt{smiler.json} file is shown in Listing~\ref{lst:AIM_smiler.json}.

As can be seen, each file contains information providing both the SMILER shortened designation for the model (in this case, \texttt{AIM}) as well as its full name and citation information. \texttt{model\_type} allows the code to easily check whether pre-requisites are available to execute the code (for example, if the MATLAB engine is not installed, SMILER will skip MATLAB-based models with a warning rather than an execution error). \texttt{model\_files} provides SMILER with a list of any files required for the model execution (\eg network weights for a CNN-based model). Model-specific parameters are specified using the same system as the global parameters in \texttt{config.json}. Additionally, some models contain a \texttt{notes} field which includes human-readable information pertinent to the specific model (such as recommendations by the original model authors or additional information which may be of use to a user).

SMILER is programmed to take a flexible approach to parameter specification, populating parameter fields according to a priority order. This order is, from greatest to least precedence: user specified values (provided at runtime, or via YAML experiment file), model-specific default values (defined in model's \texttt{smiler.json} specification), and global default values (defined in SMILER's internal config.json).

\subsection{Overview of MATLAB Interface}
\label{subsec:MATLAB_interface}

In order to help users navigate and use the code base provided by SMILER, a number of helper functions are provided. This section describes the supporting code base for the MATLAB portion of SMILER; the suite of tools which support the CLI are described in Section~\ref{subsec:SMILER_CLI}.

The primary helper file is the installation file, \texttt{iSMILER.m}. This file adds all other helper functions and all bundled MATLAB-based models to the MATLAB \texttt{path}. By default, this installation will not save the changes to the \texttt{path} beyond the current session, but a user may optionally specify that \texttt{path} changes should be permanent by calling:
\begin{lstlisting}[language=matlab, frame=none]
   iSMILER(true);
\end{lstlisting}
Should users have permanently modified the \texttt{path} and later change their mind, SMILER \texttt{path} changes may be undone by using the uninstall function provided in the file \texttt{unSMILER.m}.

The function \texttt{smiler\_info} provides a text interface in MATLAB for a user to query parameter information. This may be called without any arguments or using the string argument \texttt{'global'} to receive information about global parameters, or a specific MATLAB-based model may be specified as the input argument and the model-specific parameter and citation information for that model will be displayed.

In order to bring each included algorithm into compliance with the common API of SMILER, the code for each model is encapsulated in a wrapper function with the format \texttt{\textit{[model\_name]}\_wrap.m}, where \texttt{\textit{[model\_name]}} is a string selected from the following available list of included algorithms:
\begin{itemize}
	\setlength{\itemsep}{0cm}
	\setlength{\parskip}{0cm}
	\item{\texttt{AIM}: Attention by Information Maximization \cite{BruceTsotsos2006}}
	\item{\texttt{AWS}: Adaptive Whitening Saliency \cite{GarciaEtAl2012}}
	\item{\texttt{CAS}: Context Aware Saliency \cite{GofermanEtAl2012}, using the implementation by \cite{TsaiChangCode}}
	\item{\texttt{cG}: A centered Gaussian prior}
	\item{\texttt{CVS}: Covariance-based Saliency \cite{ErdemErdem2013}}
	\item{\texttt{DVA}: Dynamic Visual Attention \cite{HouZhang2008}}
  \item{\texttt{FES}: Fast and Efficient Saliency \cite{TavakoliEtAl2011}}
	\item{\texttt{GBVS}: Graph-Based Visual Saliency \cite{HarelEtAl2007}}
	\item{\texttt{IKN}: The Itti-Koch-Niebur Saliency Model \cite{IttiKochNiebur1998}}
  \item{\texttt{IMSIG}: Image Signature \cite{HouEtAl2012}}
  \item{\texttt{LDS}: Learning Discriminitive Subspaces \cite{FangEtAl2017}}
	\item{\texttt{QSS}: Quaternion-Based Spectral Saliency~\cite{SchauerteStiefelhagen2012}}
  \item{\texttt{RARE2012}: A multi-scale rarity-based saliency model \cite{RicheEtAl2013}}
	\item{\texttt{SSR}: Saliency Detection by Self-Resemblance~\cite{SeoMilanfar2009}}
	\item{\texttt{SUN}: Saliency Using Natural statistics \cite{ZhangEtAl2008}}
\end{itemize}

Each function operates with the following function call:
\begin{lstlisting}[language=matlab, frame=none]
output_map = [MODEL_NAME]_wrap(input_image,
                               params)
\end{lstlisting}

where the \texttt{input\_image} is either a string specifying the file path of an image or is a variable containing image data, and \texttt{output\_map} is a single-channel saliency map with the same height and width as the image specified by \texttt{input\_image}. \texttt{params} is an optional input variable in the MATLAB structure format which provides a mechanism for specifying parameter values. As mentioned in Section~\ref{subsec:Config_Files}, every model's behaviour is governed by a set of parameters which are specified as key-value pairs. If no parameter structure is provided, the wrap function will automatically populate the parameter settings with default values appropriate for the given model. If some but not all parameters are specified in the input, then the wrap function will likewise operate with default values for any unspecified structure elements.

A basic example showing the explicit calculation of four models (AIM, AWS, IKN, QSS) on an input image specified by a path string is given in Listing~\ref{lst:matlab_simple}.
\begin{lstlisting}[label={lst:matlab_simple},caption={Example MATLAB script showing the calculation of saliency maps for the AIM, AWS, IKN, and QSS models.}]
img_path = 'path/to/example.png';

AIM_map = AIM_wrap(img_path);
AWS_map = AWS_wrap(img_path);
IKN_map = IKN_wrap(img_path);
QSS_map = QSS_wrap(img_path);
\end{lstlisting}
This can be written more conveniently as a loop, iterating over the same set of models and executing each in turn and saving the saliency map as a separate image.
\begin{lstlisting}[label={lst:matlab_ex},caption={Example MATLAB script using SMILER to calculate saliency maps for the AIM, AWS, IKN, and QSS models.}]
models = {'AIM', 'AWS', 'IKN', 'QSS'};

img_path = 'path/to/example.png';

for i = 1:length(models)
  salmap = feval([models{i}, '_wrap'], img_path);
  imwrite(salmap, [models{i}, '_saliency_map.png']);
end
\end{lstlisting}
Note that the code makes use of the MATLAB \texttt{feval} function to dynamically execute code based on a string argument, which allows for a simple interface for scripting and batch execution.

Sample~\ref{lst:matlab_ex} can be easily extended if specific parameters for some models are desired. For example, if a user wanted to use one of the other learned ICA filter bases for the AIM algorithm and wanted QSS to operate over the HSV colour space (but were otherwise fine with all other default parameters), then the modified version of the script shown in Listing~\ref{lst:matlab_ex2} could be used.
\begin{lstlisting}[label={lst:matlab_ex2},caption={Example MATLAB script using SMILER to calculate saliency maps for the AIM and QSS models with customized parameters.}]
models = {'AIM', 'AWS', 'IKN', 'QSS'};

img_path = 'path/to/example.png';

for i = 1:length(models)
  params = struct();
  switch(models{i})
    case 'AIM'
      params.AIM_filters = '21infomax999.mat';
    case 'QSS'
      params.color_space = 'hsv';
  end
  salmap = feval([models{i}, '_wrap'], img_path, params);
  imwrite(salmap, [models{i}, '_saliency_map.png']);
end
\end{lstlisting} 
Note that whether the parameter is model-specific or global, the method of user specification is the same (in this example the user specifies AIM's model-specific parameter \texttt{AIM\_filters}, whereas for QSS it is the global parameter \texttt{color\_space} which is specified).

All the above samples assumes that the \texttt{iSMILER} function has already been run, and therefore all wrapper functions are available on the MATLAB \texttt{path}. Additional examples are available as part of the SMILER GitHub repository.

\subsection{Overview of SMILER CLI}
\label{subsec:SMILER_CLI}

Although the MATLAB interface is fully functional and supports all MATLAB-based models, the CLI is the recommended method of use, and future extensions to the SMILER library will likely be focused in this direction. Not only does this help migrate SMILER away from software requiring a proprietary license (MATLAB), but it also provides a more flexible platform for extension and experiment design which better supports protocol documentation.

The SMILER CLI is based on a core structure of functions which provide an interactive text-based interface to users. This includes commands to manage the containerized images for the available non-MATLAB models (see Section~\ref{subsec:Terminal_SMILER} for more details on model isolation and containerization), which at the time of this writing include the following models:
\begin{itemize}
	\setlength{\itemsep}{0cm}
	\setlength{\parskip}{0cm}
  \item{\texttt{BMS}: Boolean Map Saliency~\cite{ZhangSclaroff2016}}
  \item{\texttt{DVAP}: Deep Visual Attention Prediction~\cite{WangShen2018}}
  \item{\texttt{DGII}: DeepGaze II \cite{KummererEtAl2016}}
  \item{\texttt{eDN}: Ensemble of Deep Networks \cite{VigEtAl2014}}
  \item{\texttt{ICF}: Intensity Contrast Features \cite{KummererEtAl2017}}
  \item{\texttt{MLNet}: Deep Multi-Level Network \cite{CorniaEtAl2016a}}
	\item{\texttt{oSALICON}: Open-source Saliency in Context \cite{Thomas2016}, based on the original model by \cite{HuangEtAl2015}}
  \item{\texttt{SAM}: Saliency Attentive Model \cite{CorniaEtAl2016}}
  \item{\texttt{SalGAN}: Saliency using Generative Adversarial Networks \cite{PanEtAl2017}}
\end{itemize}

SMILER's CLI is designed to function in a Linux environment. The library is interacted with using commands with the following pattern:
\begin{lstlisting}[language=bash, frame=none]
   smiler COMMAND [OPTIONS] [ARGS]
\end{lstlisting}
where \texttt{[OPTIONS]} and \texttt{[ARGS]} are command-specific options and arguments to modify program behaviour. The SMILER commands available are as follows:
\begin{itemize}
	\setlength{\itemsep}{0cm}
	\setlength{\parskip}{0cm}
	\item{\texttt{clean}: Deletes downloaded files and docker images.}
  \item{\texttt{download}: Downloads model files and docker images.}
  \item{\texttt{info}: Provides information on SMILER models.}
  \item{\texttt{run}: Runs model(s) on images in a directory.}
  \item{\texttt{shell}: Runs a shell interface appropriate to the model environment.}
  \item{\texttt{version}: Displays SMILER version information.}
\end{itemize}
Further information about the usage of any command can be obtained by appending the \texttt{--help} flag.

Although users may directly use the CLI to conduct experiments and generate saliency maps with SMILER, the CLI additionally supports experiment specification using YAML. This is the recommended method of operation, as it allows a user to maintain explicit records of experimental settings and protocols through stored YAML specification files.

YAML is a data serialization language designed to be easily written, read, and understood by humans. SMILER uses YAML files to specify experiments. These YAML specification files are composed of two sections: an \texttt{experiment}, which provides global specification details, and one or more experimental \texttt{runs}, which provide details for a specific algorithm call. An example is presented in Listing~\ref{lst:YAML_ex}.

\begin{lstlisting}[label={lst:YAML_ex},caption={An example YAML specification file}]
experiment:
  name: Example 1
  description: An illustrative example of how to set up SMILER YAML experiments.
  input_path: /tmp/test_in
  base_output_path: /tmp/test_out
  parameters:
    do_smoothing: none

runs:
  - algorithm: AIM
    output_path: /tmp/AIM_smoothing
    parameters:
      do_smoothing: default

  - algorithm: AIM
    output_path: /tmp/AIM_no_smoothing

  - algorithm: DGII

  - algorithm: oSALICON
    parameters:
      color_space: LAB
\end{lstlisting}

The \texttt{name} and \texttt{description} fields are primarily for user records, and facilitate organization and sharing of experimental protocols by providing a lightweight document which can easily be created and stored for each experiment conducted and run on any system with SMILER installed. \texttt{input\_path} is the folder which contains the images to be processed in this particular experiment. \texttt{base\_output\_path} provides a root location for output maps to be saved, which by default will be placed in a subfolder at this location named for the algorithm that produced it (\eg in listing~\ref{lst:YAML_ex} DGII and oSALICON will be saved in \texttt{/tmp/test\_out/DGII} and \texttt{/tmp/test\_out/oSALICON} respectively).

YAML specification introduces an additional layer to parameter precedence. The \texttt{parameters} field within the \texttt{experiment} field provides a way to set customized values which will be used for all runs, but these may be overridden for a specific run by adding a \texttt{parameters} field to that run. This is demonstrated in the example shown in Listing~\ref{lst:YAML_ex}; all runs are set to be performed without smoothing based on the parameter specification under the \texttt{experiment} field, but the first run using AIM overrides this specification and instead uses default smoothing parameters. In this case, both AIM runs are include \texttt{output\_path} fields which will override the default behaviour using \texttt{base\_output\_path}. In the provided example, DGII will be run without any additional specifications beyond those provided in the \texttt{experiment} field, while oSALICON will be run with an additional specification of the \texttt{color\_space} parameter (since there is no \texttt{color\_space} specification under \texttt{experiment}, all other runs will use the built-in SMILER default: RGB).

\section{Further Implementation Details and Requirements}
\label{sec:Methods}

\subsection{MATLAB Implementation}
\label{sec:MATLAB_imp}

In addition to the interface functions mentioned in Section~\ref{subsec:MATLAB_interface}, a number of support functions are provided which are used internally by either the function wrappers or user-level helper functions (such as image loading functions which will work with either a path specification or an array variable containing image data). These functions are primarily intended for SMILER's internal use, and therefore users should not expect to interact with them directly. This includes the \texttt{jsonlab} toolbox \cite{jsonlab} for interacting with the configuration files.

We recommend using \texttt{2016a}-\texttt{2017b} versions of MATLAB. Other versions may not fully support all available MATLAB models or SMILER code (for example, the pre-compiled AWS model, for which no source code is available, does not function in newer versions of MATLAB due to the deprecation of the \texttt{princomp} function).

\subsection{Command Line Interface (CLI)}
\label{subsec:Terminal_SMILER}

As mentioned in Section~\ref{sec:Intro}, deep learning libraries are not always compatible on the same system, which presents a challenge for executing deep learning-based saliency models which rely on incompatible libraries. To solve this issue, SMILER makes use of \emph{containerization}, which is also known as \emph{operating system level virtualization}. This method creates isolated user-space instances called \emph{containers} that share the same OS kernel and drivers, but are otherwise separated. Thus, each model may be fully encapsulated within its own container, isolating any system-level libraries which may interfere with those used by other model implementations. In addition to granting this isolation, the container may be designed to provide a full specification of a model's software requirements which will be downloaded and installed upon container instantiation without the necessity of user input. This alleviates the (sometimes significant) challenge of installing all required dependencies for a given model, and allows any model encapsulated in this way to be called using a common format, analogous to the functional wrapping described in Section~\ref{subsec:MATLAB_interface}.

SMILER accomplishes containerization using Docker \cite{Docker}, and its extension \texttt{nvidia-docker} \cite{NVIDIA-Docker} which supports GPU computing. The only other dependencies for using the SMILER CLI are Python and the \texttt{click} and \texttt{yaml} Python modules.

Note that a GPU is required to efficiently run the deep learning-based models supported by SMILER. It may be possible to run all models with a compatible graphics card with at least 4GB of memory, though it is highly recommended that one use a system with 6GB or more of GPU memory.

It should be mentioned that, although all efforts have been made to eliminate code duplication in order to avoid implementation drift between the MATLAB and Python code bases, there are some support functions in the SMILER MATLAB suite that had to be re-implemented in Python in order for the CLI portion of SMILER to be able to operate independently of a MATLAB license. To ensure that these processing steps are equivalent, we maintain a set of unit tests that can be used to ensure these processing steps remain equivalent in face of future improvements to the SMILER code or new MATLAB versions.

\section{Discussion and Future Directions}
\label{sec:Discussion}

We have presented here an overview of the SMILER software package, which provides an open, standardized, and extensible framework for maintaining and executing computational saliency models. The contributions of SMILER are two-fold: a drastic reduction in human effort to set up and run saliency algorithms, and an improvement in the consistency and procedural correctness of results and conclusions produced by different research parties. SMILER is implemented and provided as an open source software project, and it is intended to foster a collaborative research community among researchers interested in exploring computational models of visual salience.

As a continually developing project, it is recommended that users familiarize themselves with SMILER documentation supplied through the GitHub project page to be made aware of any changes or updates not reflected in this document. We encourage researchers to contribute their own saliency models to SMILER, and have included a set of `skeleton' models in both MATLAB and dockerized container formats to provide a template and guidance for contributors.

\FloatBarrier
{\small
\bibliographystyle{ieee}
\bibliography{SMILER_bib}

\begin{thebibliography}{10}\itemsep=-1pt

\bibitem{Docker}
{Docker Community Edition}.
\newblock https://github.com/docker/docker-ce.

\bibitem{NVIDIA-Docker}
{NVIDIA Container Runtime for Docker}.
\newblock https://github.com/NVIDIA/nvidia-docker.

\bibitem{tensorflow2015}
M.~Abadi, A.~Agarwal, P.~Barham, E.~Brevdo, Z.~Chen, C.~Citro, G.~S. Corrado,
  A.~Davis, J.~Dean, M.~Devin, S.~Ghemawat, I.~Goodfellow, A.~Harp, G.~Irving,
  M.~Isard, Y.~Jia, R.~Jozefowicz, L.~Kaiser, M.~Kudlur, J.~Levenberg,
  D.~Man\'{e}, R.~Monga, S.~Moore, D.~Murray, C.~Olah, M.~Schuster, J.~Shlens,
  B.~Steiner, I.~Sutskever, K.~Talwar, P.~Tucker, V.~Vanhoucke, V.~Vasudevan,
  F.~Vi\'{e}gas, O.~Vinyals, P.~Warden, M.~Wattenberg, M.~Wicke, Y.~Yu, and
  X.~Zheng.
\newblock {TensorFlow}: Large-scale machine learning on heterogeneous systems,
  2015.
\newblock Software available from tensorflow.org.

\bibitem{AlnoamanyBorghi2018}
Y.~Alnoamany and J.~A. Borghi.
\newblock Towards computational reproducibility: researcher perspectives on the
  use and sharing of software.
\newblock {\em PeerJ Computer Science}, 4:e163, Sept. 2018.

\bibitem{BergaOtazu2018}
D.~Berga and X.~Otazu.
\newblock A neurodynamic model of saliency prediction in {V1}.
\newblock {\em arXiv}, abs/1811.06308, 2018.

\bibitem{BoimanIrani2007}
O.~Boiman and M.~Irani.
\newblock Detecting irregularities in iimage and in video.
\newblock {\em International Journal of Computer Vision}, 2007.

\bibitem{BorjiItti2012}
A.~Borji, D.~N. Sihite, and L.~Itti.
\newblock Salient object detection: A benchmark.
\newblock In {\em European Conference on Computer Vision {(ECCV)}}, Oct 2012.

\bibitem{BorjiEtAl2013b}
A.~Borji, D.~N. Sihite, and L.~Itti.
\newblock Quantitative analysis of human-model agreement in visual saliency
  modeling: A comparative study.
\newblock {\em IEEE Transactions on Image Processing}, 22(1):55 -- 69, 2013.

\bibitem{Brainard1997}
D.~H. Brainard.
\newblock The psychophysics toolbox.
\newblock {\em Spatial Vision}, 10:433--436, 1997.

\bibitem{BruceTsotsos2006}
N.~D.~B. Bruce and J.~K. Tsotsos.
\newblock Saliency based on information maximization.
\newblock In {\em Advances in Neural Information Processing Systems {(NIPS)}},
  volume~18, pages 155--162, 2006.

\bibitem{BruceEtAl2015}
N.~D.~B. Bruce, C.~Wloka, N.~Frosst, S.~Rahman, and J.~K. Tsotsos.
\newblock On computational modeling of visual saliency: Examining what's right,
  and what's left.
\newblock {\em Vision Research}, In Press, 2015.

\bibitem{BylinskiEtAl}
Z.~Bylinskii, T.~Judd, A.~Borji, L.~Itti, F.~Durand, A.~Oliva, and A.~Torralba.
\newblock Mit saliency benchmark.
\newblock http://saliency.mit.edu/.

\bibitem{ChangItti2010}
C.-K. Chang, C.~Siagian, and L.~Itti.
\newblock Mobile robot vision navigation and localization using gist and
  saliency.
\newblock In {\em {IEEE Conference on Intelligent Robots and Systems (IROS)}},
  2010.

\bibitem{ChangEtAl2011}
K.-Y. Chang, T.-L. Liu, H.-T. Chen, and S.-H. Lai.
\newblock Fusing generic objectness and visual saliency for salient object
  detection.
\newblock In {\em {ICCV}}, 2011.

\bibitem{CorniaEtAl2016a}
M.~Cornia, L.~Baraldi, G.~Serra, and R.~Cucchiara.
\newblock A deep multi-level network for saliency prediction.
\newblock In {\em IEEE International Conference on Pattern Recognition (ICPR)},
  pages 3488--3493, 2016.

\bibitem{CorniaEtAl2016}
M.~Cornia, L.~Baraldi, G.~Serra, and R.~Cucchiara.
\newblock Predicting human eye fixations via an lstm-based saliency attentive
  model.
\newblock {\em arXiv}, abs/1611.09571, 2016.

\bibitem{DhavaleItti2003}
N.~Dhavale and L.~Itti.
\newblock Saliency-based multifoveated mpeg compression.
\newblock In {\em Signal Processing and Its Applications, 2003. Proceedings.
  Seventh International Symposium on}, volume~1, pages 229--232 vol.1, July
  2003.

\bibitem{ErdemErdem2013}
E.~Erdem and A.~Erdem.
\newblock Visual saliency estimation by nonlinearly integrating features using
  region covariances.
\newblock {\em Journal of Vision}, 13(2013):11, 2013.

\bibitem{jsonlab}
Q.~Fang.
\newblock {jsonlab toolbox}.

\bibitem{FangEtAl2017}
S.~Fang, J.~Li, Y.~Tian, T.~Huang, and X.~Chen.
\newblock Learning discriminative subspaces on random contrasts for image
  saliency analysis.
\newblock {\em IEEE {T}ransactions on {N}eural {N}etworks and {L}earning
  {S}ystems}, 28(5):1095--1108, 2017.

\bibitem{FuEtAl2018}
K.~{Fu}, J.~{Li}, H.~{Shen}, and Y.~{Tian}.
\newblock {How Drones Look: Crowdsourced Knowledge Transfer for Aerial Video
  Saliency Prediction}.
\newblock {\em arXiv}, Nov. 2018.

\bibitem{GarciaEtAl2012}
A.~Garcia-Diaz, X.~R. Fdez-Vidal, X.~M. Pardo, and R.~Dosil.
\newblock Saliency from hierarchical adaptation through decorrelation and
  variance normalization.
\newblock {\em Image and Vision Computing}, 30(1):51--64, 2012.

\bibitem{Goble2014}
C.~Goble.
\newblock Better software, better research.
\newblock {\em IEEE Internet Computing}, 18(5):4--8, Sept 2014.

\bibitem{GofermanEtAl2012}
S.~Goferman, L.~Zelnik-Manor, and A.~Tal.
\newblock Context-aware saliency detection.
\newblock {\em IEEE Transactions on Pattern Analysis and Machine Intelligence
  ({TPAMI})}, 34(10):1915--1926, 2012.

\bibitem{GuoZhang2010}
C.~Guo and L.~Zhang.
\newblock A novel multiresolution spatiotemporal saliency detection model and
  its applications in image and video compression.
\newblock {\em {IEEE} Transactions on Image Processing}, 19:185--198, 2010.

\bibitem{HanEtAl2018}
J.~Han, D.~Zhang, G.~Cheng, N.~Liu, and D.~Xu.
\newblock Advanced deep-learning techniques for salient and category-specific
  object detection: A survey.
\newblock {\em IEEE Signal Processing Magazine}, 35(1):84--100, Jan 2018.

\bibitem{HardingRoberston2009}
P.~Harding and N.~Roberston.
\newblock Task-based visual saliency for intelligent compression.
\newblock In {\em IEEE International Conference on Signal and Image Processing
  Applications ({ICSIPA})}, 2009.

\bibitem{HarelEtAl2007}
J.~Harel, C.~Koch, and P.~Perona.
\newblock Graph-based visual saliency.
\newblock In {\em {NIPS}}, volume~19, pages 545--552, 2007.

\bibitem{HendersonEtAl2009}
J.~M. Henderson, G.~L. Malcolm, and C.~Schandl.
\newblock Searching in the dark: Cognitive relevance drives attention in
  real-world scenes.
\newblock {\em Psychonomic Bulletin and Review}, 2009.

\bibitem{HouEtAl2012}
X.~Hou, J.~Harel, and C.~Koch.
\newblock Image signature: Highlighting sparse salient regions.
\newblock {\em {IEEE} Transactions on Pattern Analysis and Machine Intelligence
  ({TPAMI})}, 34:194--201, 2012.

\bibitem{HouZhang2008}
X.~Hou and L.~Zhang.
\newblock Dynamic visual attention: Searcing for coding length increments.
\newblock {\em Neural Information Processing Systems}, 21:681--688, 2008.

\bibitem{HuangEtAl2015}
X.~Huang, C.~Shen, X.~Boix, and Q.~Zhao.
\newblock {SALICON}: Reducing the semantic gap in saliency prediction by
  adapting deep neural networks.
\newblock In {\em {ICCV}}, 2015.

\bibitem{IttiKochNiebur1998}
L.~Itti, C.~Koch, and E.~Niebur.
\newblock A model of saliency-based visual attention for rapid scene analysis.
\newblock {\em {IEEE} Transactions on Pattern Analysis and Machine Intelligence
  ({TPAMI})}, 20:1254--1259, 1998.

\bibitem{jia2014caffe}
Y.~Jia, E.~Shelhamer, J.~Donahue, S.~Karayev, J.~Long, R.~Girshick,
  S.~Guadarrama, and T.~Darrell.
\newblock Caffe: Convolutional architecture for fast feature embedding.
\newblock {\em arXiv}, 2014.

\bibitem{JuddEtAl2012}
T.~Judd, F.~Durand, and A.~Torralba.
\newblock A benchmark of computational models of saliency to predict human
  fixations.
\newblock Technical report, Massachusetts Institute of Technology, 2012.

\bibitem{KummererEtAl2016}
M.~K{\"{u}}mmerer, T.~S.~A. Wallis, and M.~Bethge.
\newblock {DeepGaze II:} reading fixations from deep features trained on object
  recognition.
\newblock {\em arXiv}, abs/1610.01563, 2016.

\bibitem{KummererEtAl2017}
M.~K{\"u}mmerer, T.~S.~A. Wallis, L.~A. Gatys, and M.~Bethge.
\newblock Understanding low- and high-level contributions to fixation
  prediction.
\newblock In {\em The IEEE International Conference on Computer Vision (ICCV)},
  Oct 2017.

\bibitem{LinEtAl2013}
J.~Y. Lin, T.~J. Liu, W.~Lin, and C.~C.~J. Kuo.
\newblock Visual-saliency-enhanced image quality assessment indices.
\newblock In {\em Signal and Information Processing Association Annual Summit
  and Conference (APSIPA), 2013 Asia-Pacific}, pages 1--4, Oct 2013.

\bibitem{MaZhang2008}
Q.~Ma and L.~Zhang.
\newblock Saliency-based image quality assessment criterion.
\newblock In {\em Proceedings of the 4th International Conference on
  Intelligent Computing: Advanced Intelligent Computing Theories and
  Applications - with Aspects of Theoretical and Methodological Issues}, ICIC
  '08, pages 1124--1133, Berlin, Heidelberg, 2008. Springer-Verlag.

\bibitem{MahadevanVasconcelos2010}
V.~Mahadevan and N.~Vasconcelos.
\newblock Spatiotemporal saliency in dynamic scenes.
\newblock {\em IEEE Transactions on Pattern Analysis and Machine Intelligence},
  32(1):171--177, Jan 2010.

\bibitem{MancasEtAl2007}
M.~Mancas, D.~Unay, B.~Gosselin, and B.~Macq.
\newblock Computational attention for defect localization.
\newblock In {\em Proc. of ICVS Workshop on Computational Attention and
  Applications}, 2007.

\bibitem{MaratEtAl2009}
S.~Marat, T.~Ho�Phuoc, L.~Granjon, N.~Guyader, D.~Pellerin, and
  A.~Gu{\'e}rin-Dugu{\'e}.
\newblock Modelling spatio-temporal saliency to predict gaze direction for
  short videos.
\newblock {\em International Journal of Computer Vision}, 82(3):231--243, 2009.

\bibitem{MasciocchiStill2013}
C.~M. Masciocchi and J.~D. Still.
\newblock Alternatives to eye tracking for predicting stimulus-driven
  attentional selection within interfaces.
\newblock {\em Human-Computer Interaction}, 28(5):417--441, 2013.

\bibitem{NuthmannHenderson2010}
A.~Nuthmann and J.~M. Henderson.
\newblock Object-based attentional selection in scene viewing.
\newblock {\em Journal of Vision}, 10(8):20, 2010.

\bibitem{PanEtAl2017}
J.~Pan, C.~Canton{-}Ferrer, K.~McGuinness, N.~E. O'Connor, J.~Torres,
  E.~Sayrol, and X.~{Gir{\'{o}} i Nieto}.
\newblock Salgan: Visual saliency prediction with generative adversarial
  networks.
\newblock {\em arXiv}, abs/1701.01081, 2017.

\bibitem{RasouliTsotsos2014}
A.~Rasouli and J.~K. Tsotsos.
\newblock Visual saliency improves autonomous visual search.
\newblock In {\em Canadian Conference on Computer and Robot Vision (CRV)},
  2014.

\bibitem{RicheEtAl2013}
N.~Riche, M.~Mancas, M.~Duvinage, M.~Mibulumukini, B.~Gosselin, and T.~Dutoit.
\newblock {RARE2012}: A multi-scale rarity-based saliency detection with its
  comparative statistical analysis.
\newblock {\em Signal Processing: Image Communication}, 28(6):642--658, 2013.

\bibitem{RobertsEtAl2012}
R.~Roberts, D.-N. Ta, J.~Straub, K.~Ok, and F.~Dellaert.
\newblock Saliency detection and model-based tracking: A two part vision system
  for small robot navigation in forested environments.
\newblock In {\em The Proceedings of {SPIE} 8387}, 2012.

\bibitem{SchauerteStiefelhagen2012}
B.~Schauerte and R.~Stiefelhagen.
\newblock Quaternion-based spectral saliency detection for eye fixation
  prediction.
\newblock In {\em European Conference on Computer Vision {(ECCV)}}, pages
  116--129, 2012.

\bibitem{SeoMilanfar2009}
H.~J. Seo and P.~Milanfar.
\newblock Static and space-time visual saliency detection by self-resemblance.
\newblock {\em Journal of vision}, 9(12):15--15, 2009.

\bibitem{SuhEtAl2003}
B.~Suh, H.~Ling, B.~B. Bederson, and D.~W. Jacobs.
\newblock Automatic thumbnail cropping and its effectiveness.
\newblock In {\em Proceedings of the 16th Annual ACM Symposium on User
  Interface Software and Technology}, UIST '03, pages 95--104, New York, NY,
  USA, 2003. ACM.

\bibitem{TavakoliEtAl2011}
H.~R. Tavakoli, E.~Rahtu, and J.~Heikkil{\"a}.
\newblock Fast and efficient saliency detection using sparse sampling and
  kernel density estimation.
\newblock In {\em Proceedings of Scandinavian Conference on Image Analysis
  {(SCIA)}}, 2011.

\bibitem{Thomas2016}
C.~L. Thomas.
\newblock {OpenSalicon}: An open source implementation of the salicon saliency
  model.
\newblock Technical Report TR-2016-02, University of Pittsburgh, 2016.

\bibitem{TsaiChangCode}
J.-F. Tsai and K.-J. Chang.
\newblock Opensource implementation of context-aware saliency detection.
\newblock
  \url{https://sites.google.com/a/jyunfan.co.cc/site/opensource-1/contextsaliency}.

\bibitem{VigEtAl2014}
E.~Vig, M.~Dorr, and D.~Cox.
\newblock Large-scale optimization of hierarchical features for saliency
  prediction in natural images.
\newblock In {\em Proceedings of the IEEE Conference on Computer Vision and
  Pattern Recognition (CVPR)}, pages 2798--2805, 2014.

\bibitem{WangShen2018}
W.~Wang and J.~Shen.
\newblock Deep visual attention prediction.
\newblock {\em {IEEE Transactions on Image Processing}}, 27(5):2368--2378,
  2018.

\bibitem{WhiteEtAl2017}
B.~J. White, J.~Y. Kan, R.~Levy, L.~Itti, and D.~P. Munoz.
\newblock Superior colliculus encodes visual saliency before the primary visual
  cortex.
\newblock {\em Proceedings of the National Academy of Sciences},
  114(35):9451--9456, 2017.

\bibitem{YuLisin2009}
S.~X. Yu and D.~A. Lisin.
\newblock {\em Advances in Visual Computing}, chapter Image Compression Based
  on Visual Saliency at Individual Scales, pages 157--166.
\newblock Springer Berlin Heidelberg, Berlin, Heidelberg, 2009.

\bibitem{YubingEtAl2010}
T.~Yubing, H.~Konik, F.~A. Cheikh, and A.~Tremeau.
\newblock Full reference image quality assessment based on saliency map
  analysis.
\newblock {\em Journal of Imaging Science and Technology}, 54, 2010.

\bibitem{ZaharescuWildes2012}
A.~Zaharescu and R.~P. Wildes.
\newblock Spatiotemporal salience via centre-surround comparison of visual
  spacetime orientations.
\newblock In {\em Asian Conference on Computer Vision ({ACCV})}, 2012.

\bibitem{ZhangSclaroff2016}
J.~Zhang and S.~Stan.
\newblock Exploiting surroundedness for saliency detection: A {B}oolean map
  approach.
\newblock {\em IEEE Transactions on Pattern Analysis and Machine Intelligence
  ({TPAMI})}, 5(38):889--902, 2016.

\bibitem{ZhangEtAl2008}
L.~Zhang, M.~H. Tong, T.~K. Marks, H.~Shan, and G.~W. Cottrell.
\newblock {SUN}: A {B}ayesian framework for saliency using natural statistics.
\newblock {\em Journal of Vision}, 8(7:32):1--20, 2008.

\bibitem{ZhuEtAl2011}
T.~Zhu, W.~Wang, P.~Liu, and Y.~Xie.
\newblock Saliency-based adaptive scaling for image retargeting.
\newblock In {\em 2011 Seventh International Conference on Computational
  Intelligence and Security}, pages 1201--1205, Dec 2011.

\end{thebibliography}
}

\begin{appendices}
\end{appendices}

\end{document}